\documentclass[12pt]{article}

\usepackage{graphicx,psfrag,epsf}
\usepackage{enumerate}
\usepackage{natbib}
\usepackage{mathtools}
\usepackage{booktabs}
\usepackage{color}
\usepackage[english]{babel} % English language/hyphenation
\usepackage[protrusion=true,expansion=true]{microtype} % Better typography
\usepackage{amsmath,amsfonts,amsthm}
\usepackage{dsfont}
\usepackage{amssymb}
\usepackage{chngcntr}
\usepackage[toc,page]{appendix}
\usepackage{textcomp}
\usepackage{url}

\usepackage{array}
\usepackage{booktabs} % Horizontal rules in tables
\usepackage{natbib}
\usepackage{setspace}

\usepackage{soul}
\usepackage{multirow}
\usepackage{enumitem}
\usepackage{microtype} % Slightly tweak font spacing for aesthetics
\usepackage[font = small,labelfont=bf,textfont=it]{subcaption}
\usepackage{xcolor}
\usepackage{tabularx}
\usepackage[font = small,labelfont=bf,textfont=it]{caption} % Custom captions under/above floats in tables or figures
\usepackage{footnote}
\usepackage{algorithm}% http://ctan.org/pkg/algorithms
\usepackage[noend]{algpseudocode}% http://ctan.org/pkg/algorithmicx
\usepackage{etoolbox}

\algblock{ParFor}{EndParFor}
\algnewcommand\algorithmicparfor{\textbf{for}}
\algnewcommand\algorithmicpardo{\textbf{do\ parallel}}
\algnewcommand\algorithmicendparfor{\textbf{end\ parallel\ for}}
\algrenewtext{ParFor}[1]{\algorithmicparfor\ #1\ \algorithmicpardo}
\algrenewtext{EndParFor}{\algorithmicendparfor}

\makeatletter
\def\BState{\State\hskip-\ALG@thistlm}

\newcommand{\distas}[1]{\mathbin{\overset{#1}{\kern\z@\sim}}}%
\newcommand{\bm}[1]{\mathbf{#1}}
\newcommand{\bb}[1]{\boldsymbol{#1}}
\newsavebox{\mybox}\newsavebox{\mysim}
\newcommand{\distras}[1]{%
  \savebox{\mybox}{\hbox{\kern3pt$\scriptstyle#1$\kern3pt}}%
  \savebox{\mysim}{\hbox{$\sim$}}%
  \mathbin{\overset{#1}{\kern\z@\resizebox{\wd\mybox}{\ht\mysim}{$\sim$}}}%
}
\newtheorem{theorem}{Theorem}
\newtheorem{proposition}[theorem]{Proposition}

\newcommand{\be}{\begin{equation}}
\newcommand{\ee}{\end{equation}}
\newcommand{\bi}{\begin{itemize}}
\newcommand{\ei}{\end{itemize}}
\newcommand{\ben}{\begin{enumerate}}
\newcommand{\een}{\end{enumerate}}

\newcommand{\D}{\mathcal{D}}
\newcommand{\E}{\mathcal{E}}

\newcolumntype{K}[1]{\geq {\centering\arraybackslash}p{#1}}
\allowdisplaybreaks

\makeatother

\let\oldbibliography\thebibliography
\renewcommand{\thebibliography}[1]{\oldbibliography{#1}
\setlength{\itemsep}{0pt}} %Reducing spacing in the bibliography.

\newcommand{\blind}{0}

\patchcmd{\footnotemark}{\stepcounter{footnote}}{\refstepcounter{footnote}}{}{}
\usepackage{booktabs}

\usepackage{makecell}
\usepackage[section]{placeins}

\begin{document}

\def\spacingset#1{\renewcommand{\baselinestretch}%
{#1}\small\normalsize} \spacingset{1}

\if1\blind
{
  \title{\bf SPlit: An Optimal Method for Data Splitting}
  \small
  \author{V. Roshan Joseph and Akhil Vakayil}\hspace{.2cm}\\
% }
  \maketitle
} \fi

\if0\blind
{
  \bigskip
  \bigskip
  \bigskip
  \begin{center}
    {\LARGE\bf  Optimal Ratio for Data Splitting}\vspace{.2cm}\\
    {V. Roshan Joseph}\vspace{.2cm}\\
    {H. Milton Stewart School of Industrial and Systems Engineering\\ 
    Georgia Institute of Technology, Atlanta, GA 30332, USA}\vspace{.2cm}\\
\end{center}
  \medskip
} \fi
%\vspace{.5in}
\bigskip

\vspace{-0.5cm}
\begin{abstract}
\noindent It is common to split a dataset into training and testing sets before fitting a statistical or machine learning model. However, there is no clear guidance on how much data should be used for training and testing. In this article we show that the optimal splitting ratio is $\sqrt{p}:1$, where $p$ is the number of parameters in a linear regression model that explains the data well. 
\end{abstract}


%\newpage
\spacingset{1.45} % DON'T change the spacing!

\section{Introduction} \label{sec:intro}
Data splitting is a commonly used approach for model validation, where we split a given dataset into two disjoint sets: training and testing. The statistical and machine learning models are then fitted on the training set and validated using the testing set. By holding out a set of data for validation separate from training, we can evaluate the performance of different models without any bias introduced during their training.

Random subsampling is the most commonly used approach for data splitting, that is, randomly sample without replacement some rows of the dataset for testing and keep the rest for training. Deterministic methods for splitting are also proposed in the literature that try to spread out the testing set so that it covers the region spanned by the original dataset in a much better way than a random testing set. CADEX \citep{Kennard1969}, DUPLEX \citep{Snee1977}, SPXY \citep{Galvao2005}, and SPlit \citep{Joseph2022} are a few examples of such deterministic methods.

The foregoing data splitting methods can be implemented once we specify a splitting ratio. A commonly used ratio is 80:20, which means 80\% of the data is for training and 20\% for testing. Other ratios such as 70:30, 60:40, and even 50:50 are also used in practice. There does not seem to be a clear guidance on what ratio is best or optimal for a given dataset. The 80:20 split draws its justification from the well-known Pareto principle, but that is again just a thumb-rule used by practitioners.

%Theoretical and numerical investigations on the optimality of data splitting ratio  so far have not lead to any consensus. \cite{picard1990data} have recommended 25-50\% for the testing set, whereas \cite{afendras2019optimality} recommended 50\%. The asymptotic analysis of \cite{larsen1999optimal} shows that this ratio should increase to 100\% as the size of the data increases, whereas the recent attempt by \cite{dubbs2021test} gives completely opposite conclusion. Numerical studies by \cite{dobbin2011optimally} have shown a preference for 33.33\% over 50\%.

Theoretical and numerical investigations on the optimality of data splitting ratio  so far have not lead to any consensus. \cite{picard1990data} have recommended 25-50\% for the testing set, whereas \cite{afendras2019optimality} recommended 50\%. The asymptotic analysis of \cite{larsen1999optimal} and \cite{dubbs2021test} show that this ratio should increase to 100\% as the size of the data increases. On the other hand, numerical studies by \cite{dobbin2011optimally} have shown a preference for 33.33\% over 50\%.

This article dwells into the the question of optimal ratio for data splitting. We propose a new criterion for evaluating the choice of splitting ratio in the next section. Based on this new criterion, we derive a  simple closed-form solution for the optimal ratio, which seems to agree with intuition and common practice. Furthermore, a practical strategy to compute the optimal ratio for a given dataset is also proposed.

\section{Optimal Ratio}
\subsection{Mathematical Formulation}
Suppose we have $N$ rows in the dataset that needs to be split into a training set of $n$ rows and testing set of $m$ rows, where $N=n+m$. Let $\gamma=m/N$ denote the splitting ratio. Our aim is to find the optimal $\gamma$ for a given dataset.

Let $\D^{train}=\{(\bm x_i, y_i)\}$, $i=1,\ldots,n$ be the training set and $\D^{test}=\{(\bm u_i, v_i)\}$, $i=1,\cdots,m$ the testing set, where the predictor variables $\bm x, \bm u\in \mathbb{R}^d$. If any of the predictor variables is categorical, we assume that they are already converted to numerical variables using some coding technique \citep{Joseph2022}. Our ultimate aim is to fit a model $g(\bm x;\bb \beta)$ which is expected to approximate the conditional expectation $E(y|\bm x)$, where $\bb \beta$ is a set of unknown parameters in the model. We will use the training set for the estimation of $\bb \beta$, and then evaluate the approximation error using the testing set.

Let $L(y,g(\bm x;\hat{\bb \beta}))$ be a loss function to assess the approximation/prediction error of the estimated model $g(\bm x;\hat{\bb \beta})$ from the training set. Then the model's generalization error is given by \citep[Ch.~7]{Hastie2009}
\begin{equation}\label{eq:err}
    \E=E\{ L(y,g(\bm x;\hat{\bb \beta}))|\D^{train}\},
\end{equation}
where the expectation is taken with respect to a new realization $(\bm x,y)$.

Assume that each row in the dataset is an independent realization from a distribution. If the rows of the testing set can also be assumed to be from the same distribution, then $\E$ can be estimated using
\begin{equation}\label{eq:errtest}
    \widehat{\E}=\frac{1}{m}\sum_{i=1}^{m}L(v_i,g(\bm u_i;\hat{\bb \beta})).
\end{equation}
If random sampling is used for obtaining the testing set for a given $m$, then $var\{\widehat{\E}\}$ is of the order $\mathcal{O}(1/m)$. \cite{Joseph2022} showed that this variance can be reduced in practice to almost $\mathcal{O}(1/m^2)$ by using support points \citep{mak2018support}, which is the basis of the SPlit method.

Now let's turn to the question of what is the optimal $m$ (or $\gamma$) to use. If we just focus on the variance of $\widehat{\E}$, then a larger $m$ might be the solution. However, a larger $m$ can lead to poor model fitting and therefore, large values of $\widehat{\E}$. Thus it makes sense to split the dataset so that we have a small generalization error with minimum variability. This can be achieved by 
\begin{equation}\label{eq:cri}
    \min_{\gamma}E\{\widehat{\E}^2\},
\end{equation}
where the expectation is taken with respect to everything that is random including the training set. Since $E\{\widehat{\E}^2\}=E^2\{\widehat{\E}\}+var\{\widehat{\E}\}$, (\ref{eq:cri}) will minimize not only the variability of the generalization error, but also its mean.

\subsection{Linear Regression}
The criterion in (\ref{eq:cri}) depends on the choice of the model $g(\bm x;\hat{\bb \beta})$ and the loss function $L(\cdot,\cdot)$. To make the optimization mathematically tractable, we will consider a special case: a linear regression model with squared error loss function.

Let $\bm f(\bm x)=(f_1(\bm x),\ldots,f_p(\bm x))'$ be the set of $p$ features formed based on the $d$ predictor variables, which can include quadratic terms, interaction terms, or other basis functions. To make the notations simple, we will also include the intercept term as part of the features, that is, $f_1(\bm x)=1$. Then, the model we would like to fit is
\begin{equation}
    g(\bm x;\bb \beta)=\bm f(\bm x)'\bb \beta.
\end{equation}
Let $\bm F_x$ be the model matrix formed using the training set, that is, the $i$th row of $\bm F_x$ is given by $\bm f(\bm x_i)'$. Assume the $rank\{\bm F_x\}=p\le n$. Then by minimizing
\[\frac{1}{n}\sum_{i=1}^n\{y_i-\bm f(\bm x_i)' \bb \beta\}^2,\]
with respect to $\bb \beta$ we obtain the familiar solution
$\hat{\bb \beta}=(\bm F_x'\bm F_x)^{-1}\bm F_x'\bm y$,
where $\bm y=(y_1,\ldots,y_n)'$. Now the generalization error can be estimated from the training set as
\[\hat{\E}=\frac{1}{m}\sum_{i=1}^m\{v_i-\bm f(\bm u_i)'\hat{\bb \beta}\}^2.\]

To compute the criterion in (\ref{eq:cri}), we need to  make a few assumptions. First, assume that $\{(\bm x_i,y_i)\}_{i=1}^n$ and $\{(\bm u_i,v_i)\}_{i=1}^m$ are two independent draws from the same distribution. Note that although each row of the dataset is an independent realization from the distribution, the rows of training and testing sets can become dependent depending on how we split the dataset. Since random sampling and SPlit maintain the distribution, the independence is a reasonable assumption, but not for the other deterministic splitting procedures such as CADEX, DUPLEX, and SPXY. Second, assume that
\begin{equation}
    E(y|\bm x)=\bm f(\bm x)'\bb \beta.
\end{equation}
In reality this assumption may not be true, but in the next section we will explain how to achieve this approximately. Third, assume that $var(y|\bm x)=\sigma^2$. Then, we have the following result, whose proof is included in the Appendix.
\begin{proposition} 
Let $\bm A=\frac{n}{m}(\bm F_x'\bm F_x)^{-1}\bm F_u'\bm F_u$. Then
\begin{align}
    E\{\widehat{\E}\}&=\sigma^2\left\{1+\frac{1}{n}E_{\bm X}\{tr(\bm A)\}\right\},\label{eq:exp}\\
    var\{\widehat{\E}\}&=\sigma^4\left\{\frac{2}{m}+\frac{4}{mn}E_{\bm X}\{tr(\bm A)\}+\frac{2}{n^2}E_{\bm X}\{tr(\bm A^2)\}+\frac{1}{n^2}var_{\bm X}\{tr(\bm A)\}\right\}\label{eq:var},
\end{align}
where $E_{\bm X}$ and $var_{\bm X}$ denote the expectation and variance taken with respect to the distribution of the predictor variables.
\label{prop2}
\end{proposition}

If 
\begin{equation}\label{eq:cond}
    \frac{1}{n}\bm F_x'\bm F_x= \frac{1}{m}\bm F_u'\bm F_u,
\end{equation}
then $\bm A=\bm I_p$ and therefore
\begin{equation}\label{eq:exact}
    E\{\widehat{\E}^2\}=\sigma^4\left\{\left(1+\frac{p}{n}\right)^2+ \frac{2}{m}+\frac{4p}{mn}+\frac{2p}{n^2}\right\}.
\end{equation}
\cite{picard1990data} suggested to split the data so that the condition (\ref{eq:cond}) holds exactly and obtained the same result using a different criterion. The ``matched split'' condition in (\ref{eq:cond}) is quite reasonable and is approximately achieved in random subsampling by the law of large numbers. In fact, the approximation gets much better if we were to use SPlit, which minimizes the energy distance \citep{szekely2013energy} between the training and testing sets \citep{Vakayil2022}. The following asymptotic result based on \cite{afendras2019optimality} sidesteps the ``matched split" requirement.

\begin{proposition}
If $\lim_{n\rightarrow \infty}\frac{1}{n}\bm F_x'\bm F_x=\lim_{m\rightarrow \infty}\frac{1}{n}\bm F_u'\bm F_u=\bm \Sigma$ is finite and positive definite, then
\begin{equation}\label{eq:asym}
    E\{\widehat{\E}^2\}=\sigma^4\left\{1+\frac{2p}{N(1-\gamma)}+\frac{2}{N\gamma}\right\}+\mathcal{O}(\frac{1}{N^2}).
\end{equation}
\end{proposition}

\begin{figure}
\begin{center}
\includegraphics[width = .6\textwidth]{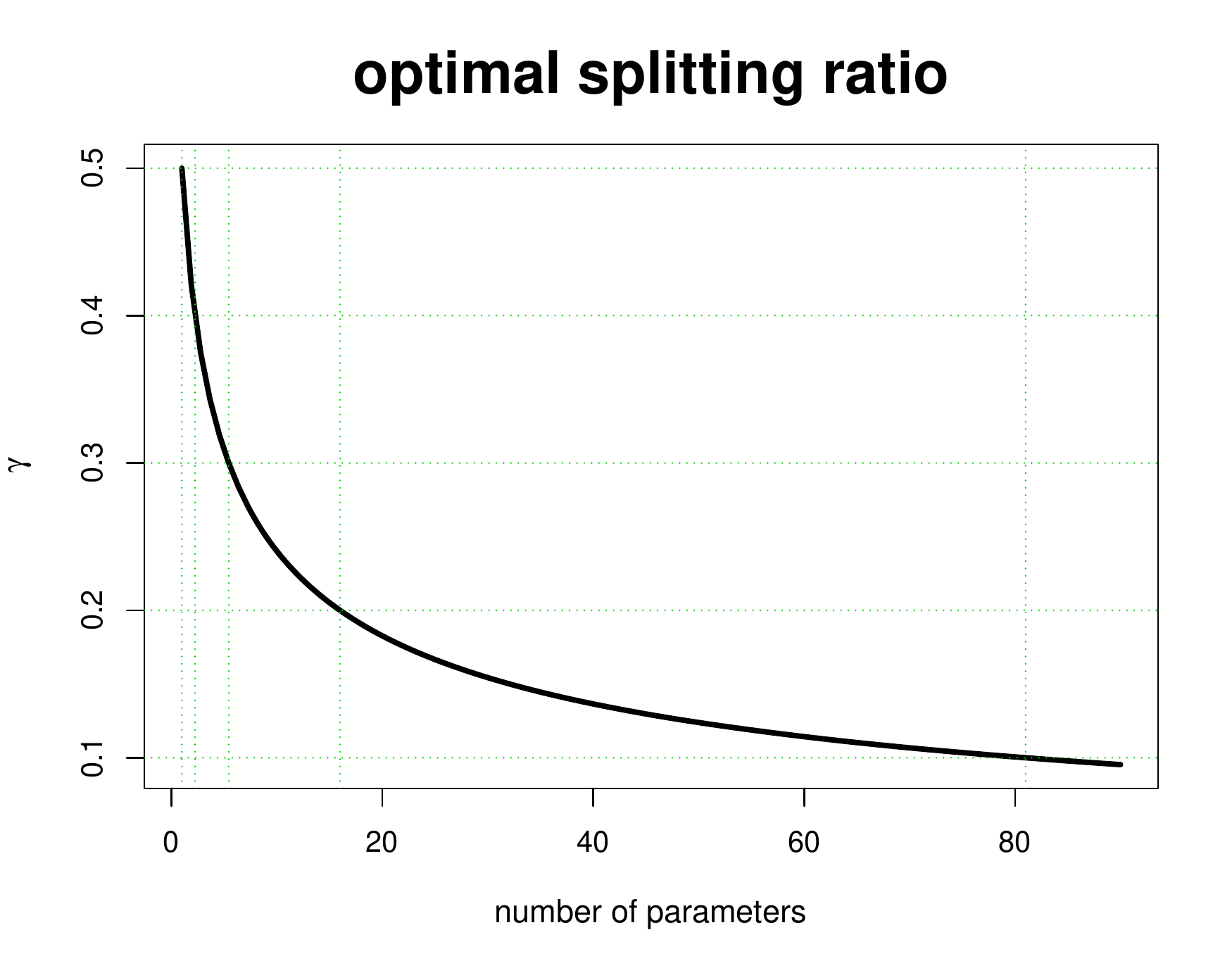} 
\caption{optimal splitting ratio against the number of parameters.}
\label{fig:optratio}
\end{center}
\end{figure}

The proof of Proposition 2 is omitted because it directly follows from Proposition 4 of \cite{afendras2019optimality}. Thus, for large $N$, the optimal splitting ratio can be obtained by minimizing
\[\frac{2p}{(1-\gamma)}+\frac{2}{\gamma}.\]
This is minimized at
\begin{equation}\label{eq:optratio}
    \gamma^*=\frac{1}{\sqrt{p}+1},
\end{equation}
which gives us the main result of this article that we should split the dataset into training and testing using the ratio $\sqrt{p}:1$. This is plotted in Figure \ref{fig:optratio}. We can see that it gives values in a range that is commonly used in practice. The curve starts at $\gamma^*=0.5$ when there is only a single parameter to estimate and then decreases to $\gamma^*=0.1$ when the number of parameters reaches 81. This behavior makes complete sense because we need more training data when there are more parameters to estimate in the model.

Sometimes it is necessary to split the training set also into two parts for estimating the regularization parameters in the model. In such a case the three sets will be called training, validation, and testing \citep[Ch.~7]{Hastie2009}. Following the optimal ratio, the split sizes for the three sets should be $\{(1-\gamma^*)^2,\gamma^*(1-\gamma^*),\gamma^*\}$. Thus, they should be split according to the ratio $p:\sqrt{p}:(\sqrt{p}+1)$. For example, if $p=16$, then the splitting ratio would be $64:16:20$.

\subsection{Simulations}
\begin{figure}
\begin{center}
\includegraphics[width = .8\textwidth, angle=90]{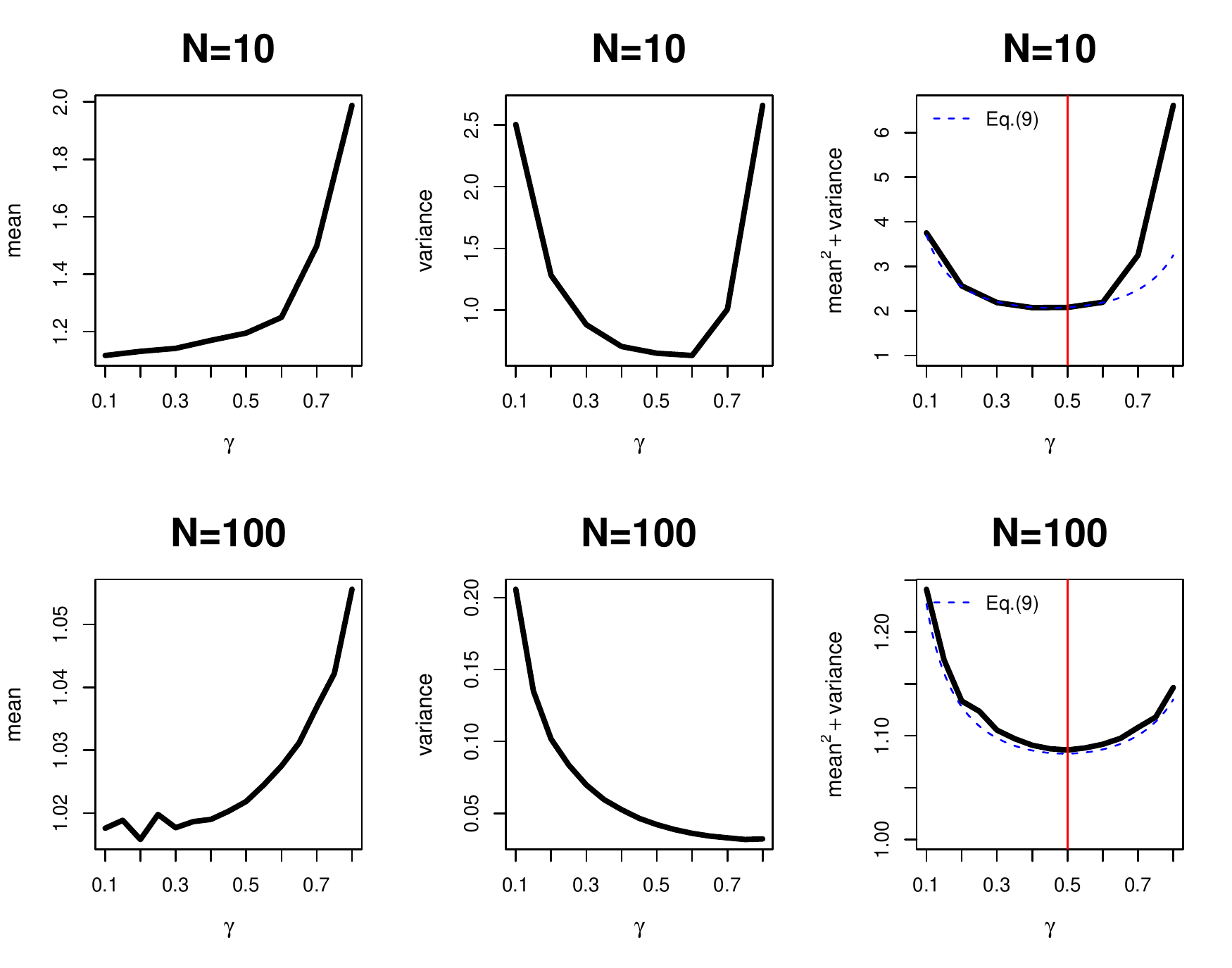} 
\caption{Plots of $E\{\widehat{\E}\}$, $var\{\widehat{\E}\}$, and $E\{\widehat{\E}^2\}$ against the splitting ratio for a model with only an intercept ($p=1$). The top panels shows the plots with $N=10$ and the bottom panels with $N=100$.}
\label{fig:sim1}
\end{center}
\end{figure}

Since several assumptions have gone into the derivation of the optimal ratio, it makes sense to check the results using simulations. First consider the simplest case with no predictor variables. We generate the data $Y_i\sim^{iid}\mathcal{N}(\mu,\sigma^2)$ for $i=1,\ldots,N$, and randomly split them into training and testing sets for a given splitting ratio $\gamma$. We then estimate $\mu$ by $\overline{y}$ from the training set and compute $\widehat{\E}=\sum_{i=1}^m(v_i-\overline{y})^2/m$ using the testing set. This is repeated 10,000 times, and the following quantities are estimated: $E\{\widehat{\E}\}$, $var\{\widehat{\E}\}$, and $E\{\widehat{\E}^2\}$. They are plotted in Figure \ref{fig:sim1} for different values of $\gamma$ and for two cases: $N=10$ and $N=100$.

We can see from Figure \ref{fig:sim1} that $E\{\widehat{\E}\}$ increases with $\gamma$, whereas $var\{\widehat{\E}\}$ decreases and then increases. This behavior of the variance might have prompted several researchers to find the optimal ratio by simply minimizing the variance. However, such an optimum can drift to 1 as $N\rightarrow \infty$. On the other hand, the proposed criterion $E\{\widehat{\E}^2\}$ is well-behaved, which has a clear minimum at $\gamma^*=0.5$. Although (\ref{eq:optratio}) is true only asymptotically as $N\rightarrow \infty$, the approximation seems to be good even for $N$ as small as 10.

%This is not surprising because random subsampling can approximately achieve the ``matched split'' %condition. For small $N$, the approximation provided by \cite{picard1990data} seems to be slightly %better. However, as we discuss in the next section that it is sufficient to use the approximation in %(\ref{eq:asym}) in practice because of the uncertainties of knowing the true model.

\begin{figure}
\begin{center}
\includegraphics[width = .5\textwidth,angle=90]{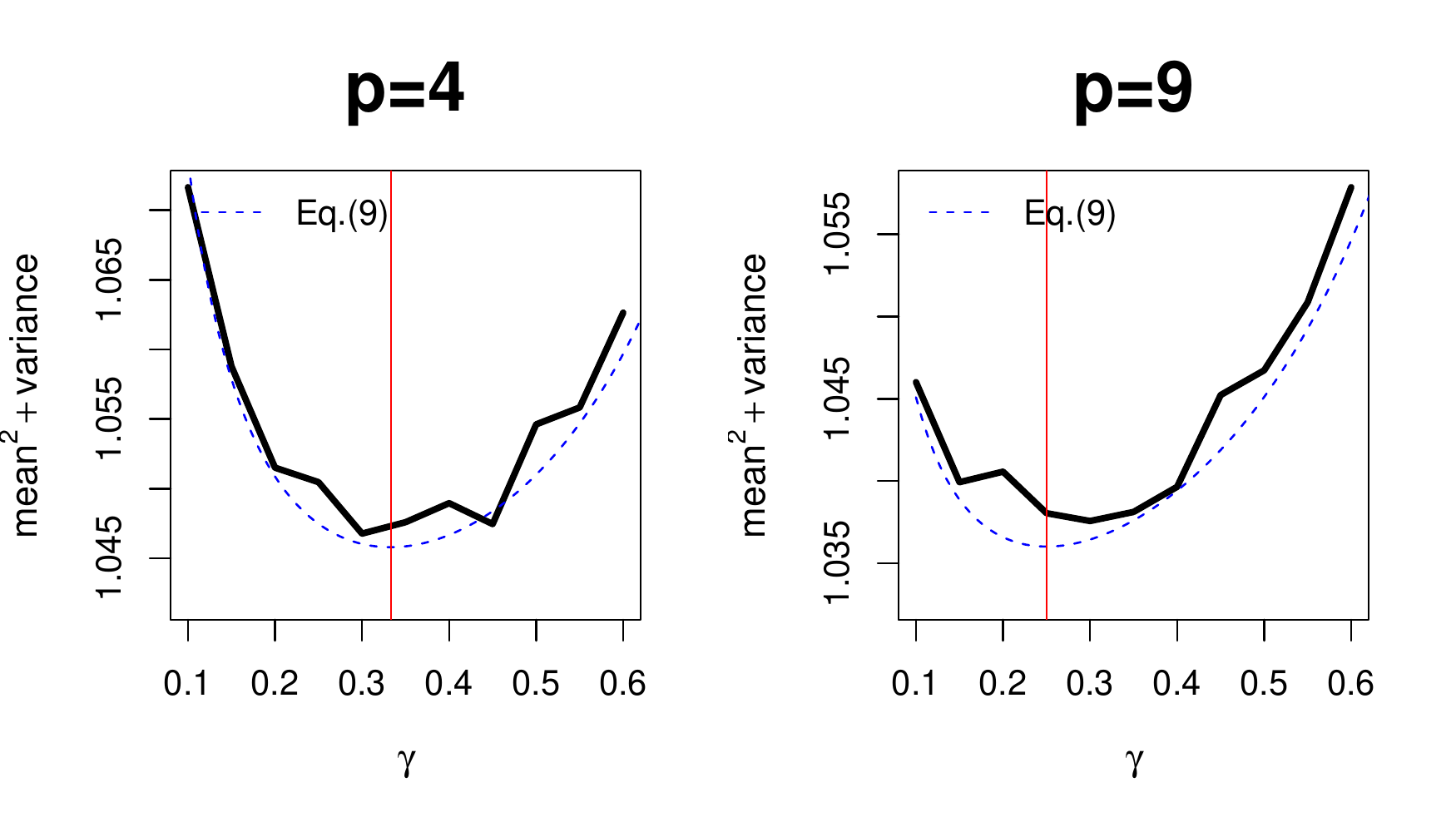} 
\caption{Plots of $E\{\widehat{\E}^2\}$ against $\gamma$ for a polynomial regression model with $p=4$ (left) and $p=9$ (right).}
\label{fig:sim2}
\end{center}
\end{figure}

Now consider a polynomial regression model with one predictor and degree $p-1$: $Y_i=\beta_0+\beta_1f_1(x_i)+\ldots+\beta_{p-1}f_{p-1}(x_i)+\epsilon_i$, where $f_r(x)$ is a Chebyshev polynomial of degree $r$ and $\epsilon_i\sim^{iid}\mathcal{N}(0,\sigma^2)$ for $i=1,\ldots,N$. Let $\{x_i\}_{i=1}^N$ be the $N$ Chebyshev nodes and $\sigma^2=1$. As before, we estimate $E\{\widehat{\E}^2\}$ for various values of $\gamma$ using 10,000 simulations with $N=100p$. Figure \ref{fig:sim2} plot two cases with $p=4$ and $p=9$. We can see that the minimum of $E\{\widehat{\E}^2\}$ is achieved around $\gamma^*$ in (\ref{eq:optratio}) up to the simulation error, confirming the validity of the theoretical result. The $E\{\widehat{\E}^2\}$ in (\ref{eq:exact}) is also plotted in Figure \ref{fig:sim2}, which agrees approximately with the simulation result. This is expected because the ``matched split''condition will be approximately achieved with random subsampling of the dataset.

\section{A Practical Strategy}
The optimal ratio in (\ref{eq:optratio}) is derived under the assumption that $E(y|\bm x)=\bm f(\bm x)'\bb\beta$. In reality, this need not be true. In fact, even if the true model is a linear regression model, we may not even know which features of the data and how many of them should be used in the model. Thus, we need a more practical strategy for deciding on a splitting ratio.

We propose a two-step approach:
\begin{enumerate}
    \item Expand the given set of predictor variables $\bm x=(x_1,\ldots,x_d)'$ into a large number of features $f_1(\bm x),\ldots,f_k(\bm x)$ and fit a linear regression model on the \emph{full} data. Use a model selection criterion such as Cp \citep{mallows1973} or AIC \citep{akaike1973} to identify the ``true'' regression model. This gives a $p\le k$.
    \item Use the $p$ identified in step 1 to compute the optimal ratio in (\ref{eq:optratio}) and split the dataset into training and testing sets.
\end{enumerate}
 
The main assumption in this approach is that the ``true model" can be well approximated by a linear regression model once we expand the feature set. We mentioned about Cp and AIC in step 2 because of its close connections to $\widehat{\E}$, but other model selection criteria can also be used. 

We will illustrate the proposed approach using a real dataset. Consider the concrete compressive strength dataset from \cite{yeh1998modeling} which can be obtained from the UCI Machine Learning Repository \citep{Dua:2019}. This dataset has eight continuous predictors pertaining to the concrete's ingredients and age, and one response: concrete's compressive strength. We create the feature set by including the main effects, two-factor interactions, and quadratic terms of the eight predictors. Stepwise regression using AIC gave a model with $p=41$ features including the intercept. Now using (\ref{eq:optratio}), we obtain $\gamma^*=0.135$. We can now split the data using this ratio. However, we may still be missing some terms in the ``true model''. For example, we may need cubic or fourth degree terms to get a good approximation. So there is a chance that the true $p$ is higher than 41. Therefore, we may take $0.135$ as an upper bound to the optimal split ratio. Thus, for this problem, it makes sense to use 0.1. In other words, 90:10 split seems like a good and reasonable choice.

\begin{figure}
\begin{center}
\includegraphics[width = .6\textwidth,angle=90]{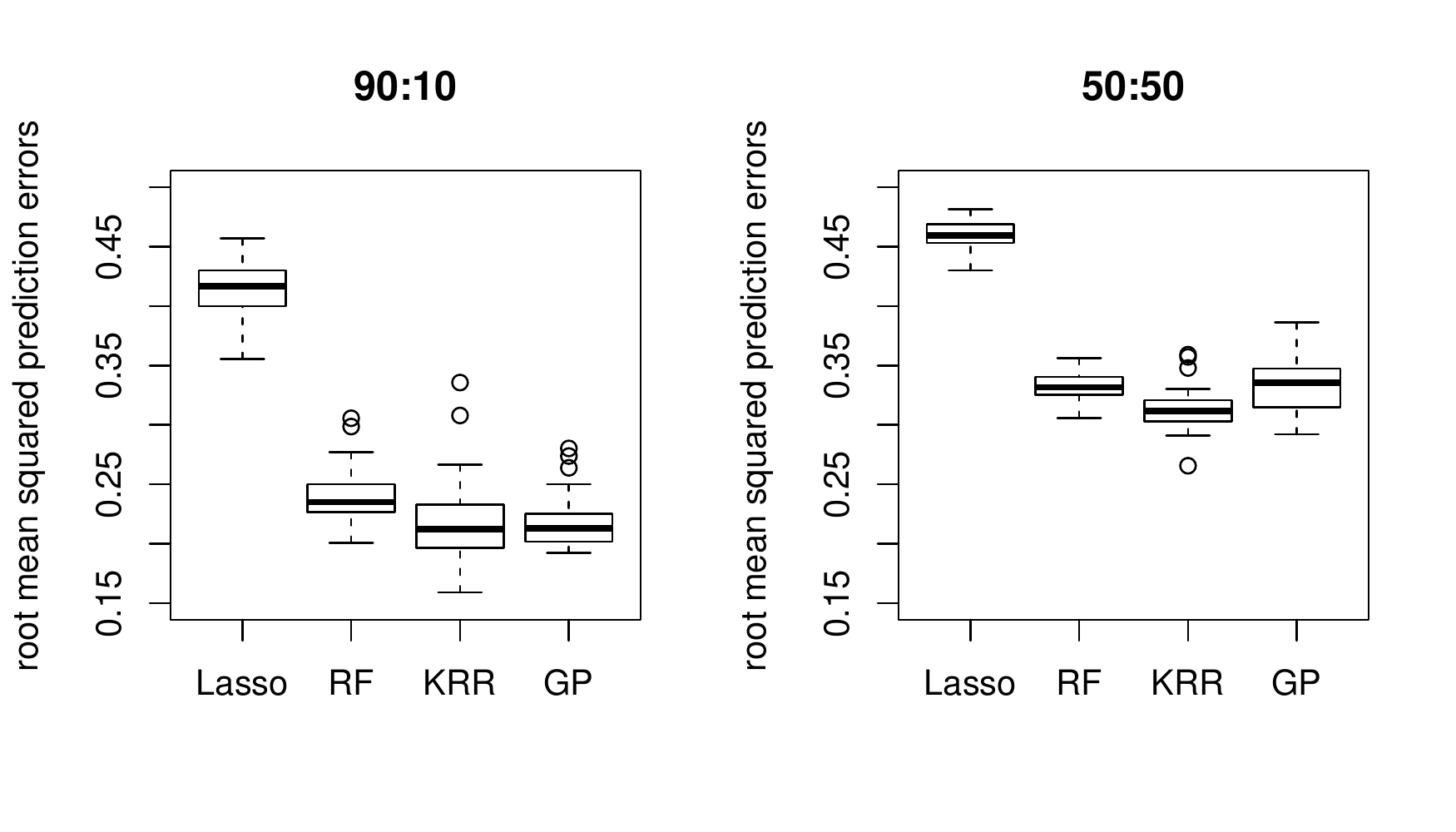} 
\caption{Boxplot of root mean squared prediction errors of different modeling methods in the concrete dataset with 90:10 split (left) and 50:50 split (right).}
\label{fig:rmse}
\end{center}
\end{figure}

To get a reliable answer with few simulations, we use SPlit \citep{Joseph2022} to split the data into training and testing sets in the ratio 90:10. Four models are fitted on the training set: (1) Lasso including quadratic and two-factor interaction terms (using R package \texttt{glmnet} \citep{glmnet}), (2) Random Forest (using R package \texttt{randomForest} \citep{randomForest}), (3) Kernel Ridge Regression (using R package \texttt{listdtr} \citep{listdtr}), and (4) Gaussian process regression (using R package \texttt{laGP} \citep{laGP}). The root mean squared prediction error  is then computed on the testing set (i.e., $\sqrt{\widehat{\E}}$). This procedure is repeated 30 times and the results are plotted in the left panel of Figure \ref{fig:rmse}.

%We can see that Kernel Ridge Regression and Gaussian process regression give the best results. In fact, both use separable Gaussian kernels and therefore, they are the same except that Kernel Ridge Regression is tuned using leave-one-out cross validation whereas Gaussian process regression is tuned using integrated likelihood. However, once we select a modeling method, we will estimate the selected model using the full data and therefore, variability across splits will not be a concern. Thus, we should pick Kernel Ridge Regression in this particular problem because it gives low prediction errors in most of the cases. The large variability of the root mean squared errors of Kernel Ridge Regression could be due to an issue with the convergence of the optimization method used for tuning the parameters. So, if special care is taken to tune it, this method should perform the best in future scenarios among the four methods considered in this study. These conclusions would not have been possible to make without using data splitting.

We can see that Kernel Ridge Regression and Gaussian process regression give the best results. In fact, both use separable Gaussian kernels and therefore, they are the same except that Kernel Ridge Regression is tuned using  cross validation whereas Gaussian process regression is tuned using integrated likelihood. However, once we select a modeling method, we will estimate the selected model using the full data and therefore, variability across splits will not be a concern. Thus, we should pick Kernel Ridge Regression in this particular problem because it gives low prediction errors in most of the cases. The large variability of the root mean squared errors of Kernel Ridge Regression could be due to an issue with the convergence of the optimization method used for tuning the parameters. So, if special care is taken to tune it, this method should perform the best in future scenarios among the four methods considered in this study. These conclusions would not have been possible to make without using data splitting.

The large errors observed for lasso compared to the other three methods indicate that the quadratic model with interaction terms is not enough to capture the ``true model''. So probably we should have included the cubic or higher order terms. This is a difficulty with linear regression models. Another approach would be to fit a nonparametric regression model on the full data and use the notion of effective number of parameters \citep{moody1991effective} to find $p$. However, since the theoretical results presented in Section 2 doesn't support this extension, we will leave this as a topic for future research.

For comparison, we have also shown the results with a 50:50 split in the right panel of the same figure. We can see that the root mean squared prediction errors of all the four methods are larger in this case, but our selection of the winner will not be affected as the Kernel Ridge Regression still seem to be doing better than the other three. However, the random forest now seems to be better than Gaussian process regression, which was not the case before. Thus, the ranking of the methods can  change with the splitting ratio and therefore, it would be better to examine the plot at the optimal splitting ratio to arrive at a more reliable result.

\section{Conclusions}
In this article we have shown that a dataset should be split in the ratio $\sqrt{p}:1$ for creating training and testing sets, where $p$ is the number of parameters to estimate in the ``true'' linear regression model. We have also discussed a practical strategy to find $p$  using model selection methods on the full dataset. 
%If one is interested in only nonparametric regression and machine learning models, $p$ could be %viewed as the effective number of parameters to estimate.

In this new era of data science, people are fitting models with millions of parameters, which might suggest that we should keep all the data for training. However, even if millions of parameters are present in a model, they are estimated with high amount of regularization, and therefore the effective number of parameters could be small. Thus the strategy given here for finding optimal data splitting ratio could still work.

Another scenario that one might encounter is the availability of physics-based models where the models may contain only a few parameters such as rate constants in a chemical kinetics model. The linear regression-based strategy proposed here might suggest a much larger $p$ as many basis functions might be needed to fully capture the response surface. However, physics-based models are derived under some simplifying assumptions and therefore, to fully validate such models we might need to estimate its discrepancy using nonparametric \citep{kennedy2001} or parametric \citep{joseph2009statistical} regression models. Thus, the strategy introduced here based on linear regression methodology is still applicable.

\vspace{.2in}
\noindent{\Large\bf Acknowledgments}

\noindent This research is supported by U.S. National Science Foundation grants and CMMI-1921646 and DMREF-1921873.

\section*{Appendix: Proof of Proposition 1}
Let $\bm X$ be the data corresponding to the predictor variables in the dataset. Since, $E(\widehat{\bb\beta}|\bm X)=\bb\beta$ and $var(\widehat{\bb\beta}|\bm X)=\sigma^2(\bm F_x'\bm F_x)^{-1}$
\begin{align*}
    E(\widehat{\E}|\bm X)&=\frac{1}{m}\sum_{i=1}^m E\{(v_i-f(\bm u_i)'\widehat{\bb \beta})^2|\bm X\}\\
    &=\frac{1}{m}\sum_{i=1}^m \{(f(\bm u_i)'\bb\beta-f(\bm u_i)'\bb\beta)^2+\sigma^2+\sigma^2f(\bm u_i)'(\bm F_x'\bm F_x)^{-1}f(\bm u_i)\}\\
    &=\sigma^2+\frac{1}{m}\sigma^2tr\{\bm F_u(\bm F_x'\bm F_x)^{-1}\bm F_u'\}\\
    &=\sigma^2\left\{1+\frac{1}{n}tr(\bm A)\right\}.
\end{align*}
Thus we obtain (\ref{eq:exp}) using $E(\widehat{\E})=E_{\bm X}[E(\widehat{\E}|\bm X)]$. Now consider the conditional variance:
\begin{align*}
var(\widehat{\E}|\bm X)&=E[var(\widehat{\E}|\bm X,\bm y)|\bm X]+var[E(\widehat{\E}|\bm X,\bm y)|\bm X]\\
&=\frac{1}{m^2}E[\sum_{i=1}^m var\{(v_i-f(\bm u_i)'\widehat{\bb \beta})^2|\bm X,\bm y\}]+\frac{1}{m^2} var\{\sum_{i=1}^m(\bm f(\bm u_i)'\bb\beta-\bm f(\bm u_i)'\widehat{\bb\beta})^2|\bm X\}\\
&=\frac{4\sigma^2}{m^2}E\{(\widehat{\bb \beta}-\bb\beta)'\bm F_u'\bm F_u(\widehat{\bb \beta}-\bb\beta)|\bm X\}+\frac{2\sigma^4}{m}+\frac{1}{m^2}var\{(\widehat{\bb \beta}-\bb\beta)'\bm F_u'\bm F_u(\widehat{\bb \beta}-\bb\beta)|\bm X\}\\
&=\frac{4\sigma^4}{m^2}tr(\bm F_u'\bm F_u(\bm F_x'\bm F_x)^{-1})+\frac{2\sigma^4}{m}+\frac{2\sigma^4}{m^2}tr(\bm F_u'\bm F_u(\bm F_x'\bm F_x)^{-1}\bm F_u'\bm F_u(\bm F_x'\bm F_x)^{-1}).
\end{align*}
Thus, (\ref{eq:var}) follows from the identity $var(\widehat{\E})=E_{\bm X}\{var(\widehat{\E}|\bm X)\}+var_{\bm X}\{E(\widehat{\E}|\bm X)\}$. 

\vspace{.1in}

\bibliography{bibliography}

\end{document}